%% file: arxiv.tex
\pdfoutput=1
\documentclass{article}

\usepackage{microtype}
\usepackage{graphicx}
\usepackage{wrapfig}
\usepackage{hyperref}
\usepackage{booktabs}
\usepackage{algorithmic}
\usepackage{algorithm}
\usepackage{amsmath}
\usepackage{amssymb}
\usepackage{mathtools}
\usepackage{amsthm}
\usepackage{bm}
\usepackage{enumitem}
\usepackage[affil-it]{authblk}

\usepackage[labelfont=bf,font=footnotesize]{caption}
\usepackage[format=hang,skip=1pt]{subcaption}
\usepackage[nameinlink]{cleveref}
\crefformat{equation}{(#2#1#3)}
\crefformat{figure}{Figure~#2#1#3}
\crefname{example}{Example}{Examples}
\crefname{lemma}{Lemma}{Lemmas}
\crefname{cor}{Corollary}{Corollaries}
\crefname{theorem}{Theorem}{Theorems}
\crefname{assumption}{Assumption}{Assumptions}

\theoremstyle{plain}

\theoremstyle{definition}

\theoremstyle{remark}

\usepackage{thm-restate}

\input{preamble/minimalist_biblatex}
\title{Does Editing Provide Evidence for Localization?\footnote{Blogpost version to appear in ICLR Blogposts 2025}}
\date{}
\author[1]{Zihao Wang}
\author[1,2]{Victor Veitch}
\affil[1]{Department of Statistics, University of Chicago}
\affil[2]{Data Science Institute, University of Chicago}

\begin{document}
\maketitle

\begin{abstract}
  A basic aspiration for interpretability research in large language models is to ``localize'' semantically meaningful behaviors to particular components within the LLM. There are various heuristics for finding candidate locations within the LLM. 
  Once a candidate localization is found, it can be assessed by editing the internal representations at the corresponding localization and checking whether this induces model behavior that is consistent with the semantic interpretation of the localization.
  The question we address here is: how strong is the evidence provided by such edits? 
  To evaluate the localization claim, we want to assess the effect of the optimal intervention at a particular location.
  The key new technical tool is a way of adapting LLM alignment techniques to find such optimal localized edits.
  With this tool in hand, we give an example where the edit-based evidence for localization appears strong, but where localization clearly fails. Indeed, we find that optimal edits at \emph{random} localizations can be as effective as aligning the full model. In aggregate, our results suggest that merely observing that localized edits induce targeted changes in behavior provides little to no evidence that these locations actually encode the target behavior.
  \end{abstract}

\section{Introduction}
A basic goal of interpretability research for large language models is to map semantically meaningful behavior to particular subcomponents of the model. 
Semantically meaningful encompasses a wide range of things, e.g., ``when asked for directions to the Eiffel Tower, the model gives directions to Paris'', ``the model responds truthfully'', or ``the model will refuse to respond''.
The aim is to find, e.g., neurons, circuits, or regions of representation space that control these behaviors.
If we could find such localizations, we could use them as building blocks to understand complex model behaviors.
Many interpretability approaches can be understood in terms of the following idealized template \cite[e.g.,][]{zou2023representation, arditi2024refusal, wang2023backdoor, chen2024truth, wei2024assessing, li2024inference, meng2022locating, vig2020causal, geiger2021causal, soulos2019discovering, finlayson2021causal, wang2022interpretability, chan2022causal, hanna2024does, conmy2023towards, todd2023function, hendel2023context}:
\begin{enumerate}
  \item We use some heuristic to find a candidate location in the model that is conjectured to be responsible for a particular behavior.
  \item We then run the model with some set of inputs, and collect the model's internal representations for each input.
  \item Then, we edit each of these representations at the candidate location, and generate new outputs according to the edited representations.
  \item If the edit changes the model's behavior in the manner that would be expected from changing the target behavior, we take this as evidence in support of localization.    
\end{enumerate}
For example, if editing a particular location in the network shifts the model to give truthful answers, we may take this as evidence that the location meaningfully encodes truthfulness in some sense. Or, if editing a location causes the model to act as though the Eiffel Tower is in Rome, we may take this as evidence that the location encodes the concept of the Eiffel Tower.
The basic question in this paper is: how strong is this evidence? That is, to what extent can we conclude that a particular location in the model is responsible for a particular behavior based on the success of editing at that location?

Our core contribution is an example where editing-based evidence appears very strong, but where localization clearly fails.
The example replicates the setup of Inference-Time-Interference (ITI) \cite{li2024inference}, where the target concept is truthfulness, and the localization is in a small subset of 16 attention heads.
Following ITI, we use logit-linear probing to identify candidate heads. 
We then search for the optimal localized edit to apply at these heads. 
Remarkably, we find that the optimal edit induces truthfulness behavior that is essentially as good as finetuning the entire model to be truthful.
That is, the localized edit is as effective as can possibly be expected.
Intuitvely, this appears to be strong evidence that the locations found by the heuristic (probing) are indeed closely linked to the target concept (truthfulness).
However, we then show that this evidence is misleading.
We find that applying optimal edits to \emph{random heads} are just as effective as when applied to the localized heads.
Accordingly, the edit-based evidence provides no support for the localization hypothesis.

A possible out here is that 16 attention heads is too many, leaving us with significant leeway to induce any behavior we want with editing.
We further strengthen the example by showing that it is possible to find a \emph{single} head in the model where editing at that head is as effective as finetuning the entire model. This appears to be the strongest edit-based evidence for localization possible.
However, we show that there are in fact multiple such heads. That is, there is simply no single privileged location that can be identified as responsible for the target behavior.

Our results suggest that the evidence provided by editing is weak, and that the success of editing at a particular location is not a reliable indicator of the location's importance for the target behavior. This seems to significantly constrain what can be learned from interpretability methods. It also points to the need for a more rigorous development of such techniques, including both precise statements of what the goals are, and well-grounded standards for evidence that these goals have been met.

The technical development in this paper relies on finding the optimal intervention at a specified location.
To that end, we develop a method for localizing LoRA type finetuning to specific locations. This then allows us to frame the search for optimal edits as a finetuning-type optimization problem. This method may also be of independent interest.

\section{Background and results from ITI}
We replicate the setup of ITI \cite{li2024inference}. 

\paragraph*{Dataset and Model Architecture}
We use TruthfulQA \cite{lin2021truthfulqa} as our dataset. It contains 817 questions that humans might answer incorrectly due to misconceptions. 
Each question contains an average of $3.2$ truthful answers and $4.1$ false answers. We use $60\%$ of the questions for training, and the rest for validation and testing. 

We use an Alpaca-7B \cite{taori2023alpaca} model that is finetuned from the Llama-7B base model. 
The model consists of $L = 32$ layers, each consisting of a Multi-head Attention (MHA) layer, and a Multilayer Perceptron (MLP) layer.
We focus on the MHA layer, which has $H = 32$ attention heads, with each head having dimension $H = 128$ (the hidden dimension is $D H = 4096$).

Ignoring MLP and layer normalization, the computation at layer $l$ can be written as:
\begin{align}
  \bm{o}_h^l &:= \text{Attn}_h^l(\bm{r}^l) \in \mathbb{R}^D\\
  \bm{o}^l &:= [(\bm{o}_1^l)^T, \ldots, (\bm{o}_H^l)^T]^T \in \mathbb{R}^{DH}\\
  W^l &:= [W_1^l, \ldots, W_H^l] \in \mathbb{R}^{DH \times DH} \label{eq:W_l}\\
  \bm{r}^{l+1} &:=  \bm{r}^l + W^l \bm{o} =  \bm{r}^l + \sum_{h=1}^H W_h^l \bm{o}_h \in \mathbb{R}^{DH}
\end{align}
where $r^l \in \mathbb{R}^{DH}$ is the residual stream before layer $l$, $\text{Attn}_h^l$ is the $h$-th attention module at layer $l$, with $\bm{o}_h^l$ being its output.
$\bm{o}^l$ is the concatenated head outputs. 
$W^l$ is the project-out matrix, that applies $H$ independent linear transformations to the corresponding head outputs.
Finally, $\bm{r}^{l+1}$ is residual stream output after layer $l$.  

\paragraph*{Localization and intervention using activation statistics}
To localize, we collect representations for positive and negative examples, and use probing to find where the truthfulness concept is represented.
To intervene, we find the direction best separating  activations for positive and negative examples, and apply this direction to the representation.

Each example is of the form, $(x, y, x_{\text{random}})$, concatenating a question $x$, a corresponding answer $y$, and another random question $x_{\text{random}}$. For positive examples, we use a truthful response $y = y_{+}$, and for negative examples, we use an untruthful response $y = y_{-}$.
To collect the representations, we feed the positive and negative examples through the model, and collect the activations of the attention heads, $\{\bm{o}_h^l\}_{h \in [H], l \in [L]}$,  \emph{at the last token}.

For each of the $L \times H$ head locations, we train a logistic regression probe on the $D$-dimensioanl activations to predict whether it's a positive or negative example.
Then we pick the attention heads with the highest probing accuracies as the localized heads.  

For the selected head at $(l, h)$, we find the direction $u^l_h$ that is ``best'' at separating the activations of positive and negative examples. There are several variants, but according to \cite{li2024inference}, the best option is the mass mean shift, which is the difference between the average positive and negative activations. 
Then we estimate the standard deviation of activations along the direction to be $\sigma_h^l$, and use the weighted direction $\theta_h^h := \sigma_h^l u_h^l$ as the intervention vector, which we add to the corresponding head during inference autoregressively.

More specifically, the applied intervention is:
\begin{align}\label{eq:iti_intervention}
  \bm{r}^{l+1}_{\text{ITI}} & := \bm{r}^l + W^l ( \bm{o} + \alpha \bm{\theta}^l)\\
  & = \bm{r}^{l+1}_{\text{orig}} + \alpha W^l \bm{\theta}^l = \bm{r}^{l+1}_{\text{orig}} + \alpha \sum_{h=1}^H W_h^l \bm{\theta}_h^l
\end{align}
where $\bm{\theta}_l$ is the concatenated intervention vectors across all heads at layer $l$, and $\alpha$ is the intervention strength.
This intervention is repeated for each next token prediction autoregressively until the whole answer is completed.

\paragraph*{Evaluation Metrics} Since the goal is to assess model's generation quality, it's natural to use truthfulness score and informativeness score of generations as the evaluation metrics. They use GPT-judge models \cite{lin2021truthfulqa} to evaluate the model's generations for truthfulness and informativeness, and use Info*Truth (the product of scalar truthful and informative scores) as the main metric. 

We also report other metrics as in the ITI paper: KL divergence of the model's next-token prediction distribution post- versus pre-intervention, and multiple-choice accuracy (MC) which is  determined via comparing the conditional probabilities of candidate answers given the question.

\section{Editing Localized Heads Modifies the Output as Expected}\label{sec:heuristic_results}
In ITI, the authors find that editing on $16$ localized heads (out of a total of $1024$ heads) successfully steers model generations to be more truthful while still being informative. 
They also find intervening on all attention heads doesn't make model generations more truthful than intervening just at the localized heads.
This seems to suggest that the truthfulness concept is indeed encoded in the localized heads. 

\begin{figure}[!htb]
  \centering
  \begin{subfigure}{0.32\textwidth}
      \centering
      \includegraphics[width=\textwidth]{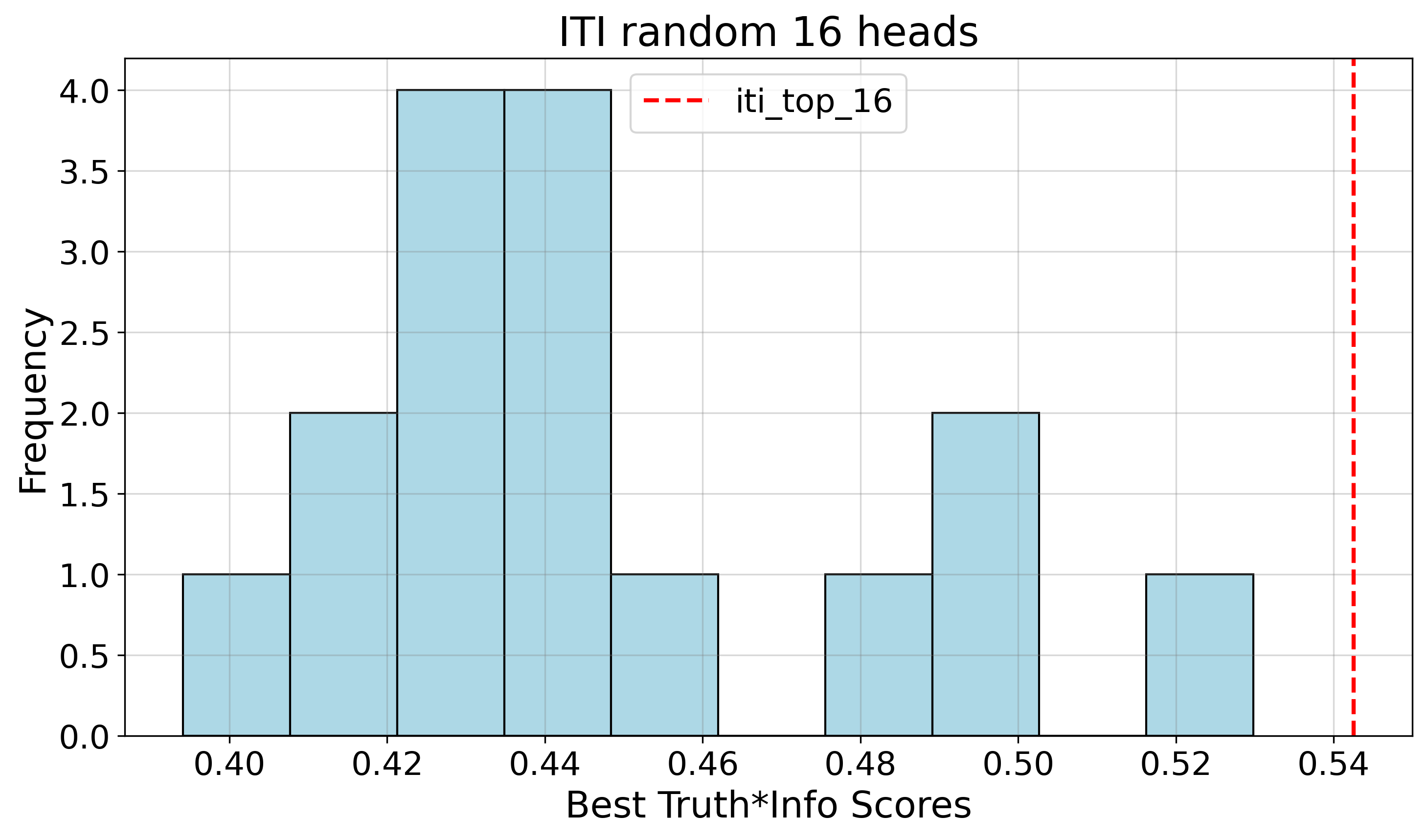}
      \caption{Info*Truth Scores}
      \label{fig:hist_iti}
  \end{subfigure}
  \hfill
  \begin{subfigure}{0.32\textwidth}
      \centering
      \includegraphics[width=\textwidth]{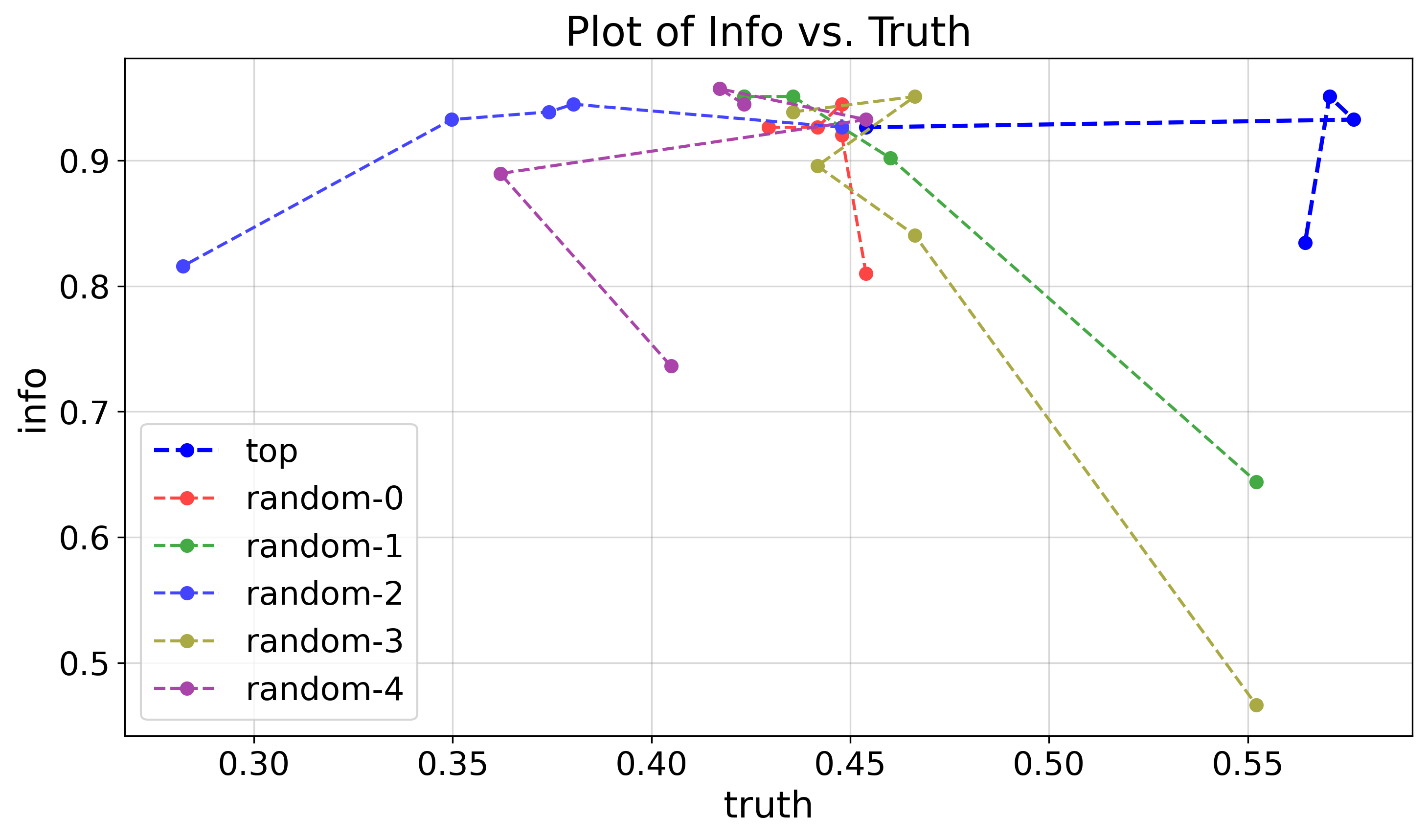}
      \caption{Truth vs Info Scores}
      \label{fig:iti_truth_info}
  \end{subfigure}
  \hfill
  \begin{subfigure}{0.32\textwidth}
      \centering
      \includegraphics[width=\textwidth]{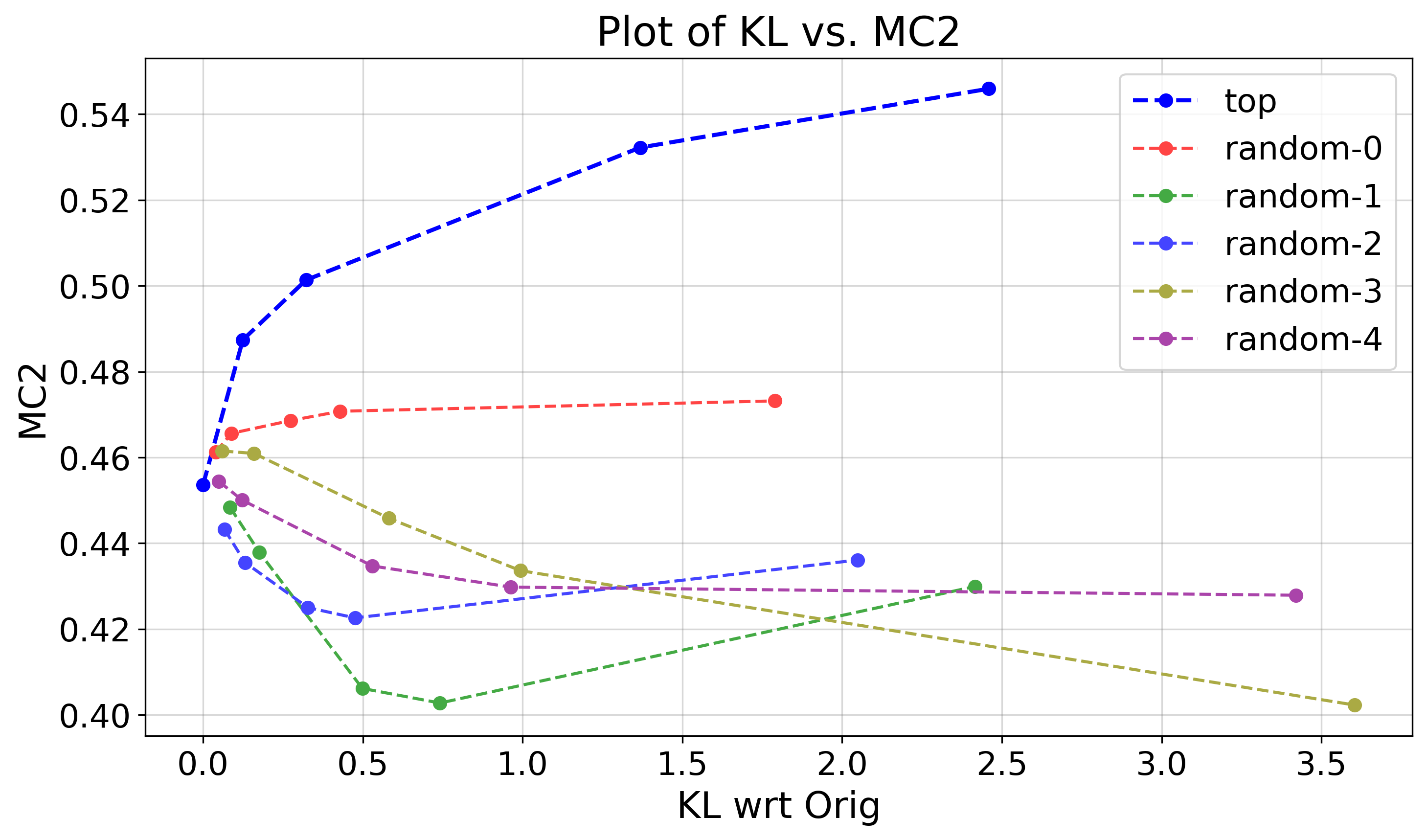}
      \caption{KL vs MC Scores}
      \label{fig:iti_kl_mc}
  \end{subfigure}
  \caption{Localized heads perform much better than random when using ITI interventions. We observe better Info*Truth scores, better truth-info score tradeoff, as well as better MC-KL tradeoff.}
  \label{fig:evidence_for_loc}
\end{figure}

We now strengthen this evidence further.
Similar to \Cite{hase2024does}, we check if interventions at random heads can also make model generations more truthful. 
More specifically, 
\begin{enumerate}
  \item Randomly select $16$ heads, and compute intervention vectors $\bm{\theta}^l$'s accordingly. 
  \item Apply varying intervention strength $\alpha$,  collect model generations, and compute scores for truthfulness and informativeness using GPT-judge across all intervention strengths.
  \item Repeat for $16$ times. 
\end{enumerate}
We find that interventions at the localized heads are more effective than interventions at random heads. In \cref{fig:hist_iti} we report the Info*Truth score (average truthfulness score times average informativeness score). We find that using localized heads have significantly higher Info*Truth scores than using random heads (p-value $1.6 \times 10^{-8}$). In fact, using random heads often doesn't have noticeable effect on the truthfulness at all, as shown in \cref{fig:iti_truth_info}, \cref{fig:iti_kl_mc}. 

This appears to add further evidence that the localized heads are ``special'' for the truthfulness concept.
However, this strong association could be because the intervention and localization are ``correlated'', since both use statistics of the same activations (determined by the design of the data, etc.). E.g. for heads with very low probing accuracy, the estimated intervention vectors could be very noisy, and thus the interventions could be less effective.

\section{Finding ``optimal'' interventions}
To test whether a particular behavior is localized to specific location, we would like to assess the effect of the \emph{optimal} intervention at that location. In the case of our running example, we want the localized edit to the representation space that does the best job of steering the model's generations to be more truthful while maintaining informativeness. Then, the questions are: what is the best we could hope to achieve? (I.e., what is ``optimal''?) And, (how) can we find a localized edit that achieves it?

\subsection*{Fitting the alignment objective gives optimal interventions}

\begin{figure}[!htb]
  \centering
  \includegraphics[width=0.8\textwidth]{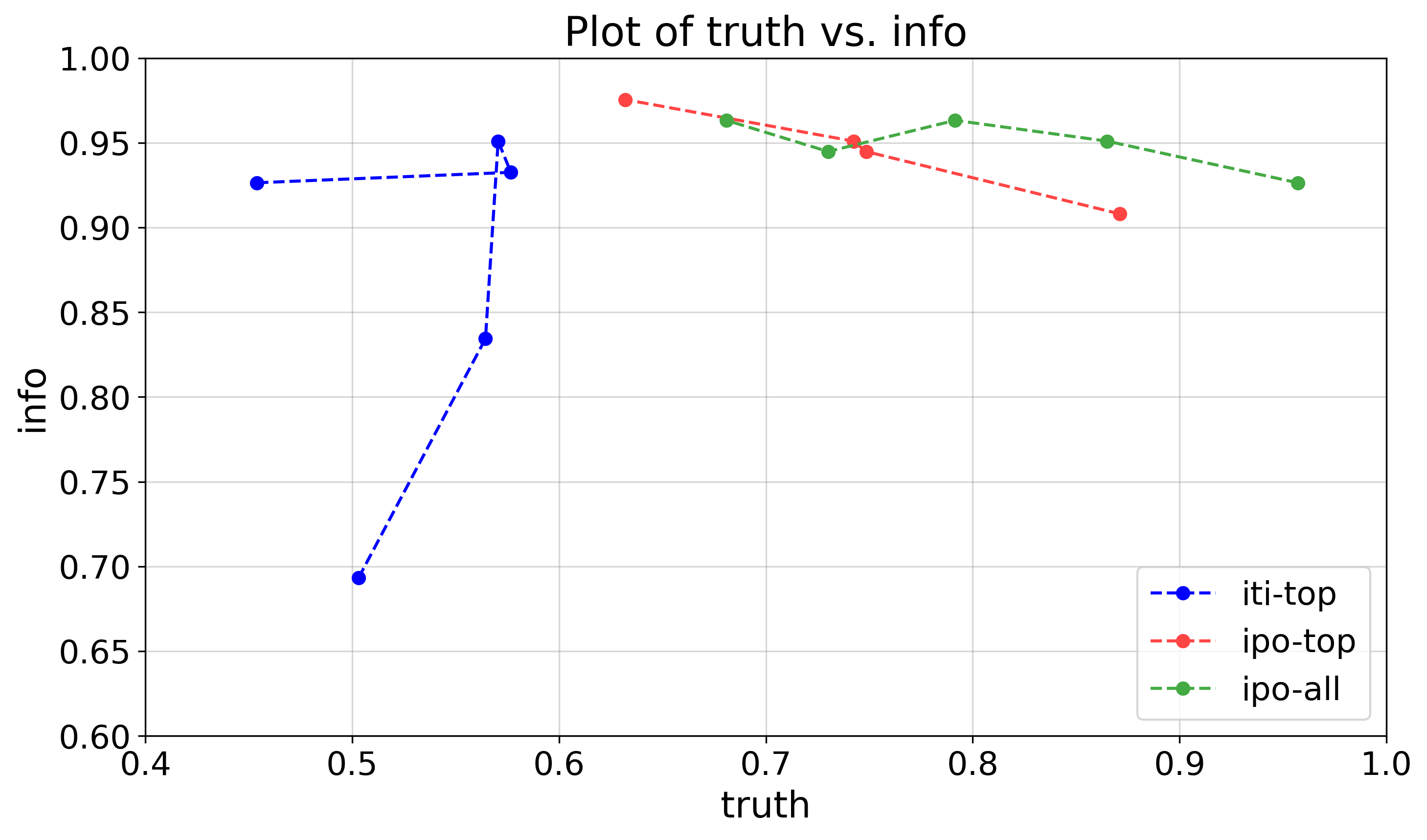}
  \caption{IPO interventions achieve much better performance than using ITI. Using IPO interventions at localized heads give nearly optimal info-truth tradeoff as well.}
  \label{fig:stronger_evidence_for_loc}
\end{figure}

The key observation is that the dataset used to construct positive and negative examples can be restructured as paired ``preference'' data $\{(x_i, y_i^{+}, y_i^{-})\}_i$, where $x_i$ is the question, $y_i^{+}$ is the truthful answer, and $y_i^{-}$ is the untruthful answer.
Since the goal is to make model generations more truthful, we can directly adopt contrastive alignment methods for biasing the model towards the truthful answers.
In this case, we use the IPO \cite{azar2024general} learning objective, where the goal is to upweight probabilities for $y_i^{+}$ and downweight probabilities for $y_i^{-}$ (up to some threshold):
\begin{align*}
  \text{argmax}_{\phi} \sum_i \left[\log \left( \frac{\pi_{\phi}(y_i^{+} | x_i)}{\pi_0(y_i^{+} | x_i)} / \frac{\pi_{\phi}(y_i^{-} | x_i)}{\pi_0(y_i^{-} | x_i)} \right) - \frac{\tau^{-1}}{2}\right]^2
\end{align*}
where $\pi_{\phi}(.|x)$ is the model's generation probability, $\pi_0(.|x)$ is the original model's generation probability, and $\tau$ decides the threshold. 
Ideally, the optimized  $\pi_{\phi^*}(.|x)$ should generate responses that are more truthful than the original model, while minimally affecting the off-target aspects of the generation (in this case, the informativeness of the responses).

To test the effectiveness of IPO alignment, we finetune the weights for project-out matrices $W^l$'s defined in \cref{eq:W_l} using (rank 1) LoRA \cite{hu2021lora}.
The finetuned model gives nearly perfect trade-off between truthfulness and informativeness, that is far better than ITI interventions \cref{fig:stronger_evidence_for_loc}.
This also suggests that ITI heuristics are very far from optimal, and contrasts with ITI results that intervening on all heads doesn't make model generations more truthful.

Now we treat this result as the overall best performance that we can achieve with interventions.
We want to see if optimal interventions at localized heads can achieve the same performance, and if random heads can achieve the same performance.

\subsection*{Connect weight updates to representation editing}
The connection to IPO lets us search for the best possible update to the model's \emph{weights}. However, we are interested in localized edits to model \emph{representations}.
To continue, we need to connect the weight editing to representation editing.

\paragraph*{Rank-1 LoRA}
Directly applying rank-1 LoRA to $W^l$, we can view the effect of adding in the modified LoRA weight matrix as an edit to the representation as follows: 
\begin{align}\label{eq:lora_edit}
  \bm{r}_{\text{LoRA}}^{l+1} := \bm{r}^l + (W^l + \bm{b}^l (\bm{a}^l)^T)\bm{o}^l = \bm{r}_{\text{orig}}^{l+1} + \left< \bm{a}^l, \bm{o}^l\right> \bm{b}^l,
\end{align}
where $\bm{a}^l, \bm{b}^l$ are the LoRA weights to optimize. 
Comparing with \cref{eq:iti_intervention}, we see that $\bm{b}^l$ plays the role of the added $W^l \bm{\theta}^l$, and $\left< \bm{a}^l, \bm{o}^l\right>$ is the intervention strength but is adapted to the representation $\bm{o}$.\footnote{One could replace $\left< \bm{a}^l, \bm{o}^l\right>$ with a constant intervention strength, but allowing the extra flexibility is closer to the ideal of best-possible-localized-intervention.}

This formulation connects weight edits to representation edits. However, it doesn't yet allow us to localize edits to specific heads --- while $\bm{\theta}^l$ can be read as concatenation of headwise intervention vectors, the projected $W^l \bm{\theta}^l$ have no corresponding interpretations. Therefore, we can't restrict the edits to specific heads by imposing structure on $\bm{b}^l$'s. 

\paragraph*{Rank-1 LoRA with reparameterization}
We can make more direct connections by reparameterizing $\bm{b}^l$ with $W^l \bm{b}^l$ (without changing expressiveness):
\begin{align}\label{eq:lora_edit_reparam}
  \bm{r}_{\text{LoRA-reparam}}^{l+1} := \bm{r}_{\text{orig}}^{l+1} + \left< \bm{a}^l, \bm{o}^l\right> W^l \bm{b}^l = \bm{r}_{\text{orig}}^{l+1} + \left< \bm{a}^l, \bm{o}^l\right> \sum_{h=1}^H W_h^l \bm{b}_h^l
\end{align} 
Here $\bm{b}_h^l$ plays the role of the intervention vector $\bm{\theta}_h^l$, and $\bm{a}^l$ decides the intervention strength adaptively.

Now we have the algorithm to find the optimal interventions for the chosen set of heads:
\begin{enumerate}
  \item Finetune the model weights using reparameterized LoRA with the IPO objective.
  \item And, restrict $\bm{b}^l$ to be nonzero only for the chosen set of heads.
\end{enumerate}

\section{Optimal interventions at localized heads are nearly optimal, but so are random heads}

\begin{figure}[!htb]
  \centering
  \begin{subfigure}{0.48\textwidth}
      \centering
      \includegraphics[width=\textwidth]{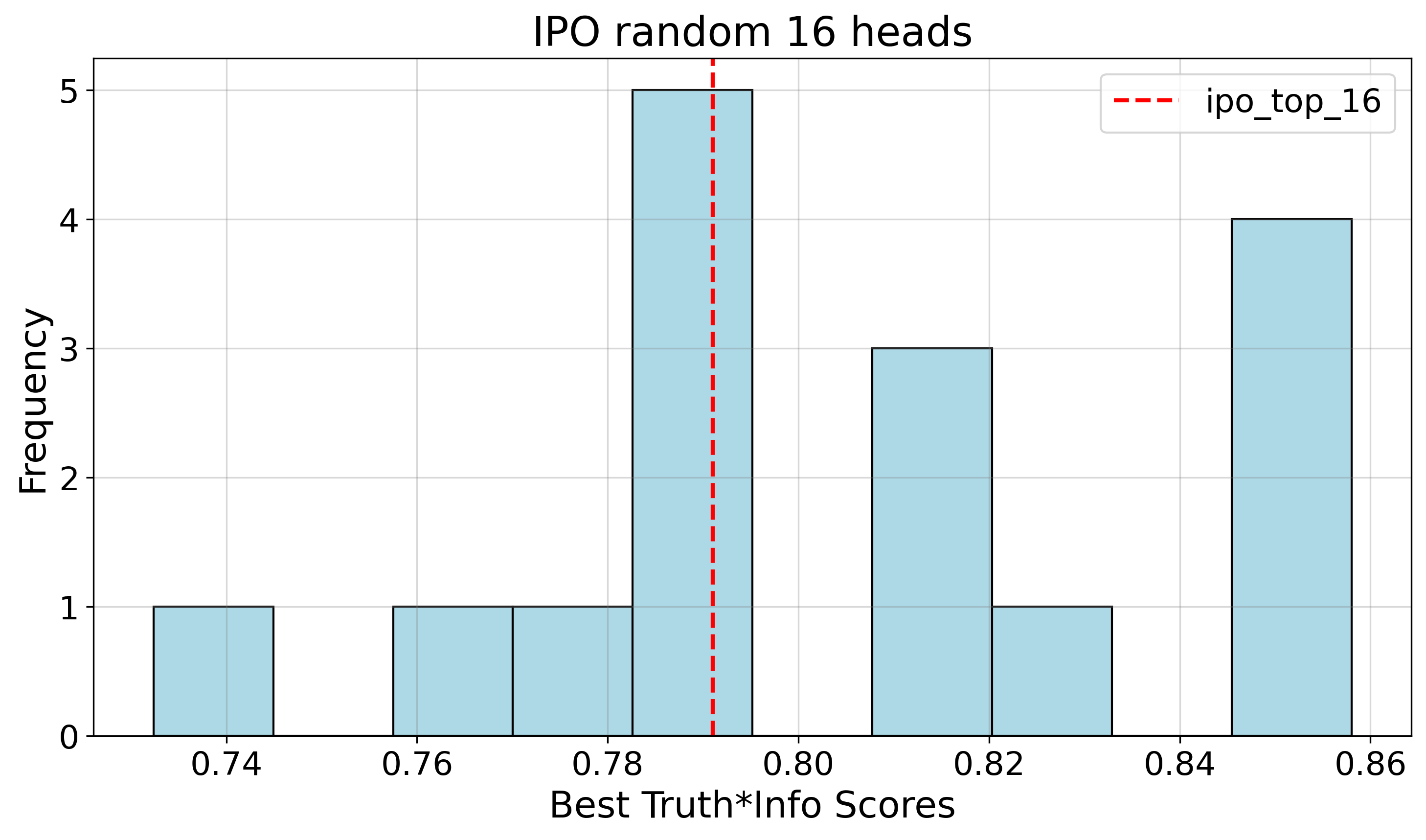}
      \caption{Info*Truth Scores}
      \label{fig:hist_ipo}
  \end{subfigure}
  \hfill
  \begin{subfigure}{0.48\textwidth}
      \centering
      \includegraphics[width=\textwidth]{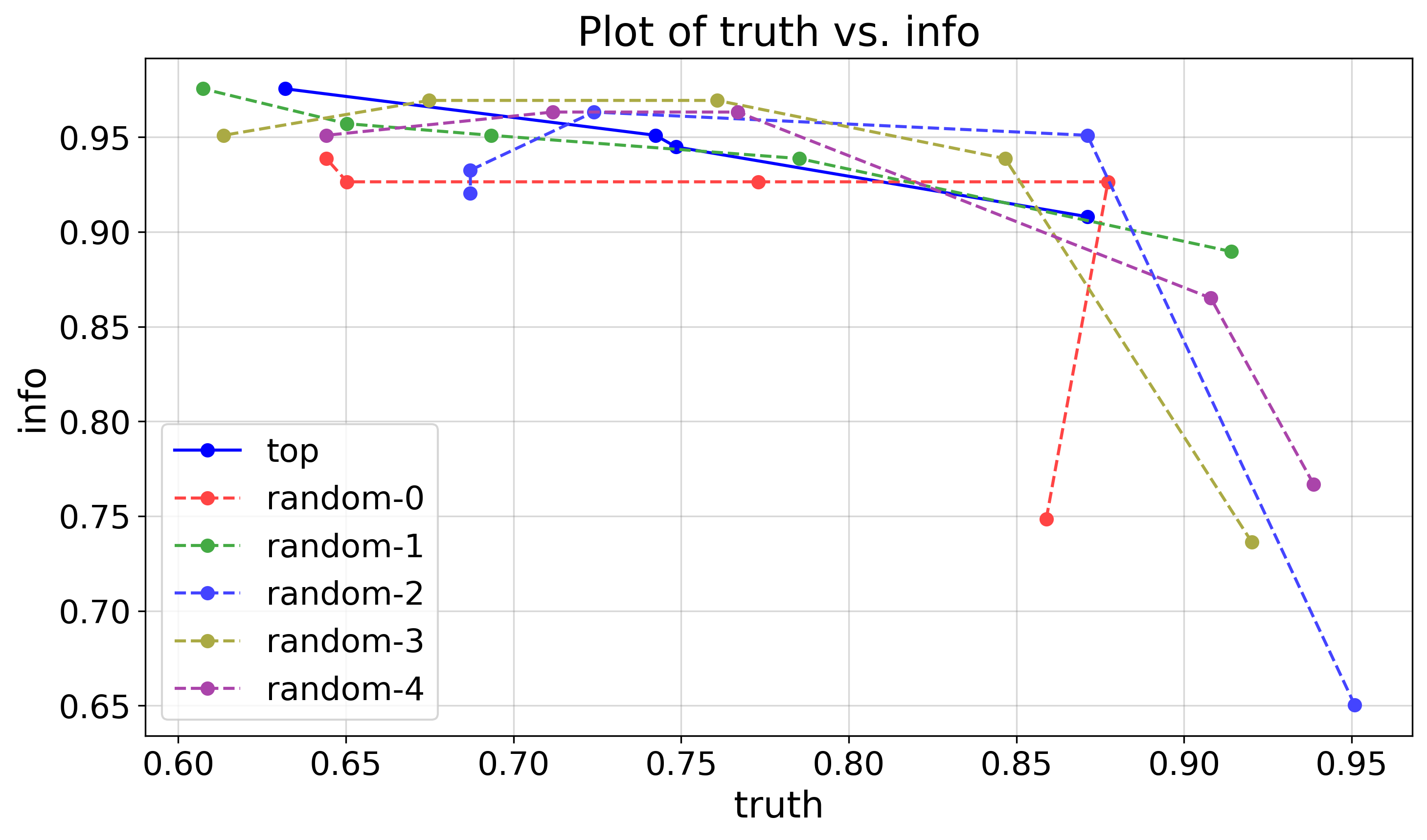}
      \caption{Truth vs Info Scores}
      \label{fig:random_vs_top}
  \end{subfigure}
  \caption{Using IPO optimal localized interventions, randomly selected heads perform nearly optimally for steering model generations. In particular, random heads are as good as the conjectured localized heads. The random heads are the same as those in \cref{fig:iti_truth_info}.}
  \label{fig:random_heads_as_good}
\end{figure}

\paragraph*{Optimal Edits at Conjectured Localization}
We can now search for the best possible interventions at the localized heads. \Cref{fig:stronger_evidence_for_loc} shows the result. We find that the optimal interventions strongly outperform the heuristic ITI interventions. Moreover, the localized interventions are about as effective as full IPO alignment! This appears to be the strongest edit-based evidence for localization that we could hope for.

\paragraph{Optimal Edits at Random Localization}
Now, we apply the same optimal edit procedure to 16 \emph{randomly selected} heads. \Cref{fig:random_heads_as_good} shows the results. In short: the optimal interventions at random heads are often just as effective as the optimal interventions at the localized heads. Accordingly, the fact that editing at the localized heads was effective at steering generations provides no evidence that the truthfulness concept is localized to those heads.

Further, the random heads we use here are the same random heads used in \cref{sec:heuristic_results}. Using the ITI heuristic intervention, the selected heads looked highly different from these random heads. But we now see that this appears to be an artifact of the suboptimal interventions and choice of metric, rather than a meaningful difference in how the heads relate to truthfulness.

\section{Intervening a single head is just as effective}

\begin{figure}[!htb]
  \centering
  \begin{subfigure}{0.48\textwidth}
      \centering
      \includegraphics[width=\textwidth]{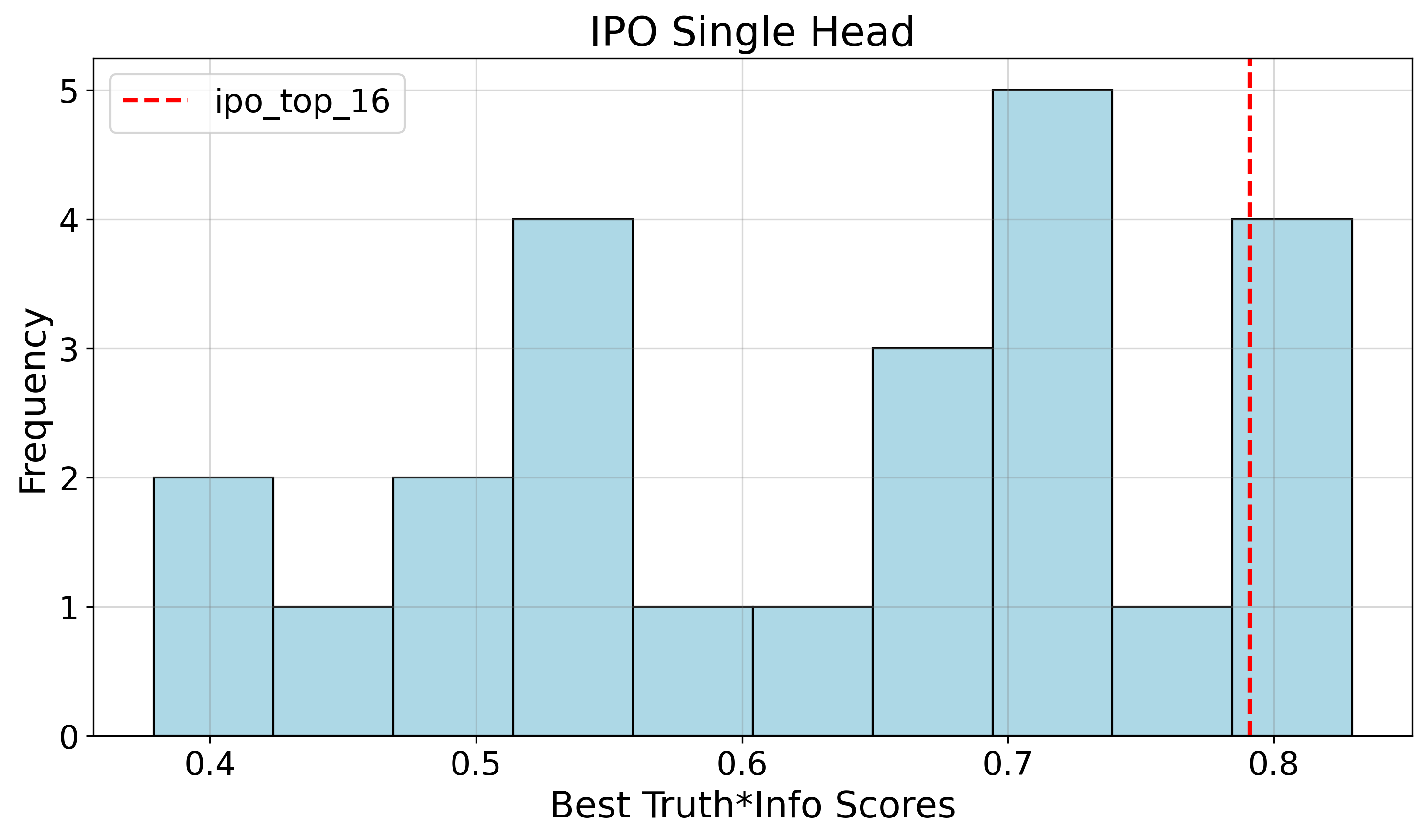}
      \caption{Info*Truth Scores}
      \label{fig:hist_ipo_1}
  \end{subfigure}
  \hfill
  \begin{subfigure}{0.48\textwidth}
      \centering
      \includegraphics[width=\textwidth]{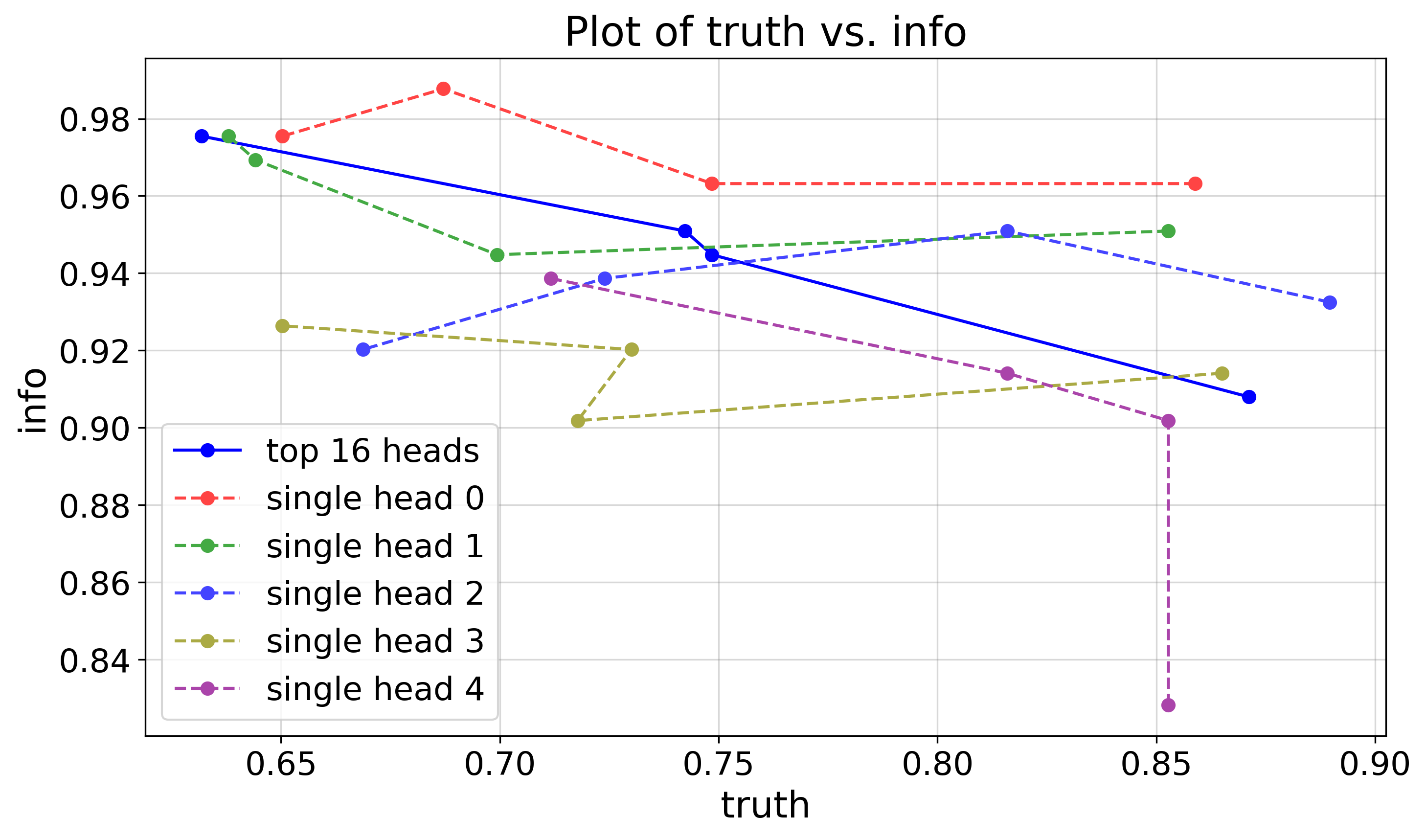}
      \caption{Truth vs Info Scores}
      \label{fig:single_vs_top}
  \end{subfigure}
  \caption{Using a single-head is as effective, and there are multiple of them!}
  \label{fig:unifying}
\end{figure}

It is now clear that edit-based evidence does not provide strong evidence for localization in the 16 head setup. However, a possible way of saving localization would be to argue that 16 heads is too many, giving too much leeway to induce any behavior we want with editing.
For example, if we edited half the heads of the model, it would not be surprising if we could make the model do anything we wanted. Accordingly, we might hope that there is still a valid syllogism of the form ``the localized edit is extremely constrained'' and ``edits at this location optimally control the target behavior'' implies ``the target behavior is localized to this location''.

To test this, we now focus on the single head case.
The procedure is simple: we randomly sample $24$ single heads, one at a time, and search for optimal interventions. The distribution of the best Info*Truth scores is shown in \cref{fig:hist_ipo_1}. 
We find $5$ single-heads that are as effective \cref{fig:single_vs_top}, and none of them has high probing accuracy. 

Notice that, still, none of these heads can be understood as localizing the truthfulness concept. The reason is that there are multiple distinct locations that work equally well! That is, even in the extreme case of a very localized edit that replicates the target behavior essentially optimally, we still cannot conclude that there is evidence supporting localization.

\section{Are the Probing-Localized Heads Anything Special?}

So far what we mean by localization, is that we can change model generation on target concept by an edit at this location. And our experiments show no evidence for this type of localization, and probing-localized heads play no special role.

So, are the probing-localized heads anything special at all?

\begin{figure}[!htb]
  \centering
  \includegraphics[width=0.8\textwidth]{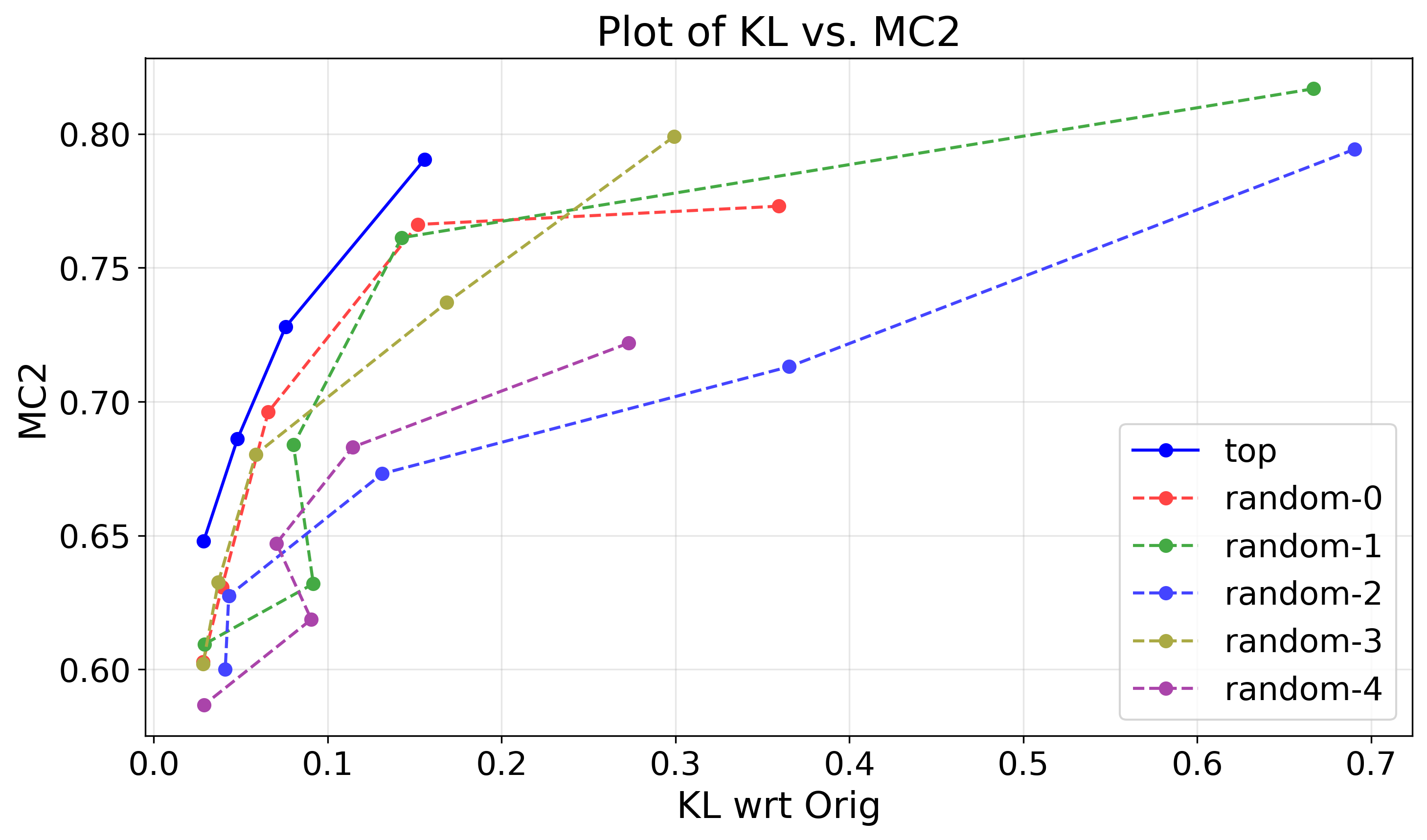}
  \caption{Probing-localized heads seem somewhat special in MC scores.}
  \label{fig:random_vs_top_MC_KL}
\end{figure}

\paragraph*{Probing-localized heads seems special for MC scores}
We do observe that these heads achieve slightly better Multiple-Choice (MC) scores compared to randomly selected heads (see \cref{fig:random_vs_top_MC_KL}), although this advantage is not as pronounced as with the ITI interventions (see \cref{fig:iti_kl_mc}). Thus, these heads may be special in terms of changing model probabilities on the given fixed dataset, which is what MC measures.

\paragraph*{The gap between what the model ``knows'' and what it generates}
It's important to note that the model's probabilities for fixed responses, do not directly correspond to what the model actually generates. 
Even if the model assigns a higher probability to a truthful response than an untruthful one, it may still not generate the truthful response if the fixed dataset is off-policy (i.e. both probabilities are low). This highlights the well-known gap between what a model ``knows'' (which is the motivation behind probing) and what it ultimately generates \cite{joshi2023personas, wang2020language, kadavath2022language, saunders2022self, burns2022discovering}.

\paragraph*{Implications}
It's possible that while probing-localized heads are not special at all for controlling model generations, they are special in changing what the model ``knows''.
Though we caution that the results here are not rigorous evidence for localization even in this sense.
Even if there is a knowledge localization in some sense, it is clear that this does not inform steering, and does not give a way of monitoring model behavior (because changes in completely unrelated locations can change the behavior). This points to the need for making the goal of localization precise.

\section{Discussion}
The main idea in this paper is that to assess the localization of a behavior we should study the effect of the \emph{optimal} intervention at the conjectured localization. The main obstacle is that, in general, it is not clear how to define or find the optimal intervention. To overcome this, we map the problem of finding the optimal intervention to the problem of finding the optimal weight update, which can be solved using existing LLM alignment methods.

The main result is an example where, naively, the evidence for localization appears strong, but when we use optimal interventions, the evidence disappears. 

The particular example---truthfulness and ITI-based evidence---was selected simply because the data used to define the heuristic happens to also allow us to set up a contrastive alignment problem. The most limited read of the results here is that ITI interventions do not provide evidence for localization, and that truthfulness does not appear to be localizable. However, the broader point is that by giving an example where editing-based evidence doesn't support localization, we see that in general such edits---by themselves---cannot provide evidence for localization. This is true irrespective of the particular behavior or heuristic being evaluated.

Thus far, we've been a bit vague about what localization means. Editing does tautological evidence for localization in the sense of ``it's possible to modify model behavior on such-and-such a behavior by an edit at this location''. On the opposite end, the strongest possible standard would be to show that the location is unique, or at least necessary. This is the standard that would be required if our aim was, e.g., to establish that LLM truthfulness can be monitored by examining a small set of heads. Potentially, there are interesting and useful notions of localization in between these two extremes. However, we can see no useful sense of localization that is consistent with the location being only as good as a \emph{randomly selected} alternative.
As we have seen, heuristic edit-based evaluation cannot even rule out this case.

Our findings add to a growing body of work that assesses the validity of interpretability results. \Cite{niu2024does} argue that the Knowledge Neuron thesis, which suggests that facts are stored in MLP weights, is an oversimplification and does not adequately explain the process of factual expression in language models. \Cite{makelov2023subspace} demonstrate that subspace activation patching can lead to an illusory sense of interpretability, as the effects may be achieved through dormant parallel pathways rather than the hypothesized subspaces. Most relevant to our work, \Cite{hase2024does} find that localization conclusions from causal tracing do not provide insight into which model MLP layer would be best to edit to override an existing stored fact. 

Overall, the results here point to the need for precise statements of what the objectives are in interpretability. With clear objectives, it may be possible to develop theoretically grounded methods for evaluation. Precise, falsifiable, statements and clear standards of evidence would suffice to prevent the kind of failure we observe in this paper.

\newpage

\printbibliography
\newpage

\appendix

\section{Experiment Details}

\paragraph*{Dataset and Model Architecture}
We use the TruthfulQA dataset \cite{lin2021truthfulqa} and the Alpaca-7B model \cite{taori2023alpaca} for our experiments. The dataset contains 817 questions with truthful and untruthful answers. We turn them into pairs, and use $60\%$ for training (6560 paired data) and the rest for validation and testing. The model consists of $32$ layers, each with $32$ attention heads and a hidden dimension of $4096$.

\paragraph*{Training Details}
We use IPO objective \cite{azar2024general} and use hyperparameter $\tau = 0.1, 0.2, 0.3, 0.4, 0.5$. We train for two epochs with a cosine scheduler, with a batch size of 4. We use ``paged\_adamw\_32bit'' optimizer.
For training with different numbers of heads, we find a smaller number of heads benefit from a higher learning rate. For all-heads, we use a learning rate of $1 \times 10^{-4}$, and for 16 heads, we use $5 \times 10^{-4}$. For single-head, we use $2 \times 10^{-3}$.

\paragraph*{Evaluation Metrics}
We reuse code from ITI \cite{li2024inference} for evaluation when possible. For GPT-judge models, we follow \cite{lin2021truthfulqa} and finetune on truthfulness and informativeness dataset using OpenAI API \cite{openai2020api}. Our finetuned model achieves similar validation error as in \cite{lin2021truthfulqa}.

\newpage

\end{document}

%% file: preamble/minimalist_biblatex.tex
\usepackage{csquotes}
\usepackage[%
minnames=1,maxnames=99,maxcitenames=2,
style=alphabetic,
doi=false,
url=false,
giveninits=true,
hyperref,
natbib,
backend=bibtex,
sorting=nyt,
backref=true
]{biblatex}%
\renewbibmacro{in:}{%
  \ifentrytype{article}{}{\printtext{\bibstring{in}\intitlepunct}}}

\renewbibmacro*{journal}{%
  \iffieldundef{journaltitle}
    {}
    {\printtext[journaltitle]{%
       \printfield[noformat]{journaltitle}%
       \setunit{\subtitlepunct}%
       \printfield[noformat]{journalsubtitle}}}}

\DeclareFieldFormat{sentencecase}{\MakeSentenceCase*{#1}}

\renewbibmacro*{title}{%
  \ifthenelse{\iffieldundef{title}\AND\iffieldundef{subtitle}}
    {}
    {\ifthenelse{\ifentrytype{article}\OR\ifentrytype{inbook}%
      \OR\ifentrytype{incollection}\OR\ifentrytype{inproceedings}%
      \OR\ifentrytype{inreference}\OR\ifentrytype{misc}}
      {\printtext[title]{%
        \printfield[sentencecase]{title}%
        \setunit{\subtitlepunct}%
        \printfield[sentencecase]{subtitle}}}%
      {\printtext[title]{%
        \printfield[titlecase]{title}%
        \setunit{\subtitlepunct}%
        \printfield[titlecase]{subtitle}}}%
     \newunit}%
  \printfield{titleaddon}}

\AtEveryBibitem{%
\ifentrytype{article}{
    \clearfield{urldate}%
    \clearfield{eprint}
    \clearfield{eid}
}{}
\ifentrytype{book}{
    \clearfield{url}%
    \clearfield{urldate}%
    \clearfield{eprint}
}{}
\ifentrytype{collection}{
    \clearfield{url}%
    \clearfield{urldate}%
    \clearfield{eprint}
}{}
\ifentrytype{incollection}{
    \clearfield{url}%
    \clearfield{urldate}%
    \clearfield{eprint}
}{}
}

\AtEveryBibitem{
    \clearfield{pages}
    \clearfield{review}%
    \clearfield{series}%
    \clearfield{volume}
    \clearfield{month}
    \clearfield{isbn}
    \clearfield{issn}
    \clearlist{location}
    \clearfield{series}
    \clearlist{publisher}
    \clearname{editor}
}{}